\pdfoutput=1 

\documentclass[a4paper, 10pt, conference]{ieeeconf}      

\IEEEoverridecommandlockouts                              

\usepackage{graphicx}
\usepackage{epsfig} 
\usepackage{mathptmx} 
\usepackage{times} 
\usepackage{amsmath} 
\usepackage{amssymb}  
\usepackage{url}
\usepackage{booktabs}

\graphicspath{ {./images/} }

\title{\LARGE \bf
Zero-shot Degree of Ill-posedness Estimation for Active Small Object Change Detection
}

\author{Koji Takeda$^{1,3}$ and Kanji Tanaka$^{2}$ and Yoshimasa Nakamura$^{1}$ and Asako Kanezaki$^{3}$ 
\thanks{$^{1}$K.Takeda and Y.Nakamura are with Tokyo Metropolitan Industrial Technology Research Institute,
         Tokyo, Japan
        {\tt\small \{takeda.koji\_1, nakamura.yoshimasa\}@iri-tokyo.jp}}%
\thanks{$^{2}$K. Tanaka is with Faculty of Engineering, University of Fukui, Japan.
        {\tt\small tnkknj@u-fukui.ac.jp}}%
\thanks{$^{3}$K. Takeda and A. Kanezaki are with School of computing, Department of Computer Science, Tokyo institute of Technology, Japan.
        }%
}

\begin{document}

\maketitle
\thispagestyle{empty}
\pagestyle{empty}

\newcommand{\subcaption}[1]{\footnotesize #1}

\newcommand{\figA}{
\begin{figure}[t]
  \centering
  \includegraphics[width=8cm]{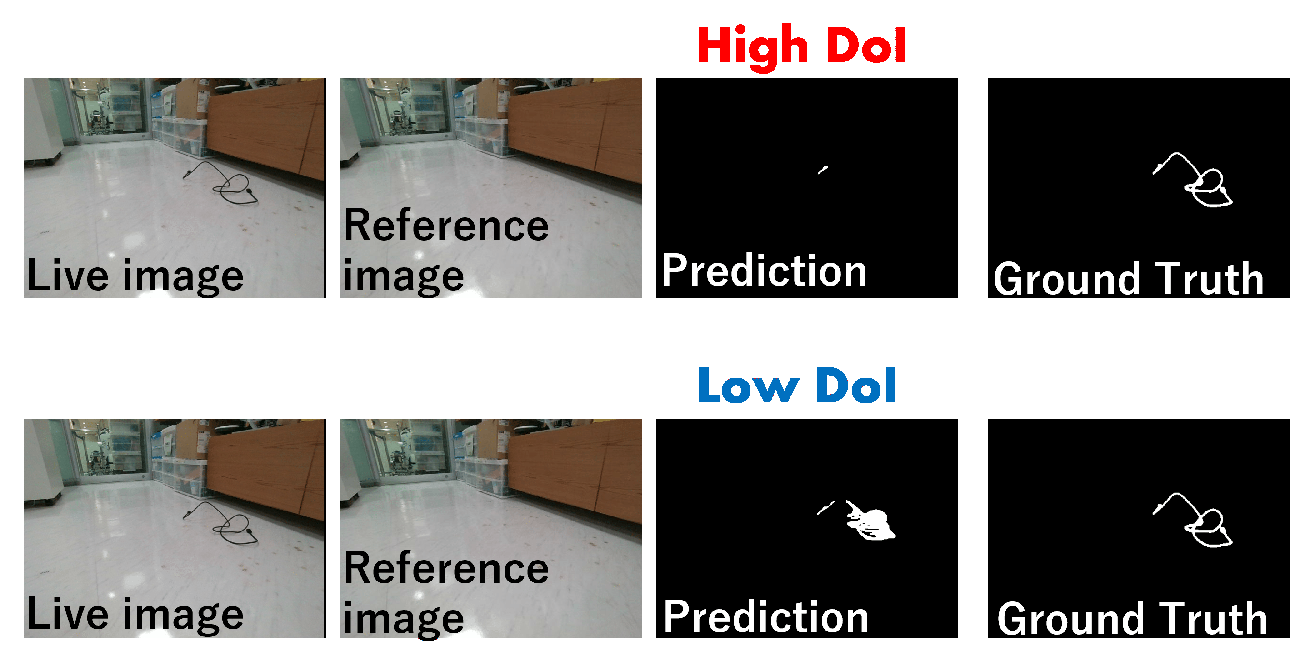}\\
\caption{%
Degree of Ill-posedness (DoI) in small object change detection. If DoI is high, robot should temporarily suspend its own navigation task and and trigger an active change-detection task in which it approaches and closely inspects potential small object changes.
}
\vspace*{-6mm}
\label{fig:A}
\end{figure}
}

\newcommand{\figB}{
\begin{figure*}[t]
  \centering
  \includegraphics[width=18cm]{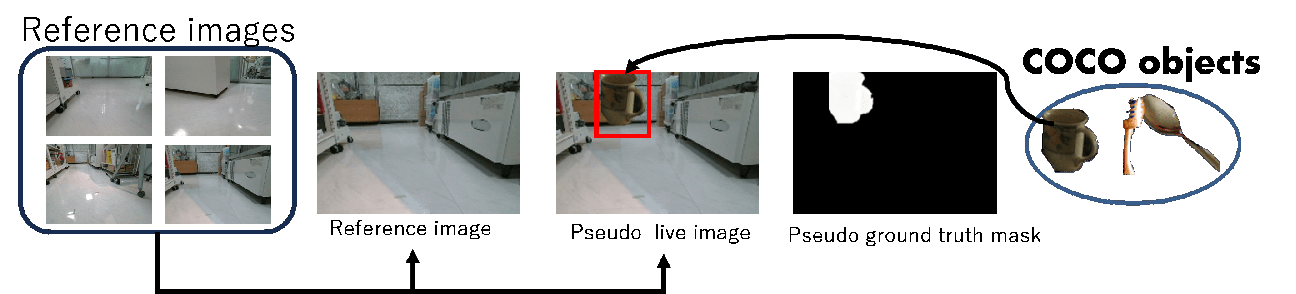}\\
\caption{%
Overview of synthetic training set creation. Randomly select an image from the reference images and paste a COCO object to obtain a reference image, pseudo live image, and pseudo ground truth mask.
}
\vspace*{-3mm}
\label{fig:B}
\end{figure*}
}

\newcommand{\figP}{
\begin{figure*}[t]
  \centering
  \includegraphics[width=17.5cm]{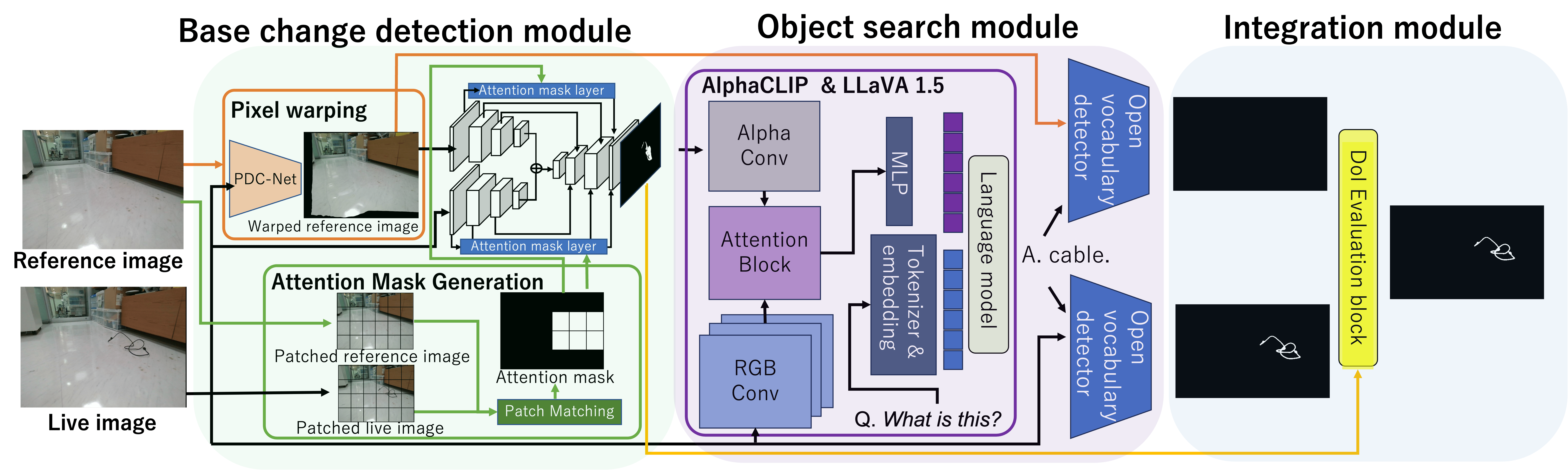}\\
\caption{%
Proposed change detection framework. First, a binary change mask is obtained from two images, a reference image and a live image, in the Base change detection module. Next, the object search module get object mask for reference image and live image by utilizing large multimodal model and open vocabulary segmentation.
Finally, the final change detection result is calculated while evaluating the DoI using the Integration module.
}
\vspace*{-6mm}
\label{fig:P}
\end{figure*}
}

\newcommand{\figC}{
\begin{figure}[t]
  \centering
  \includegraphics[width=8cm]{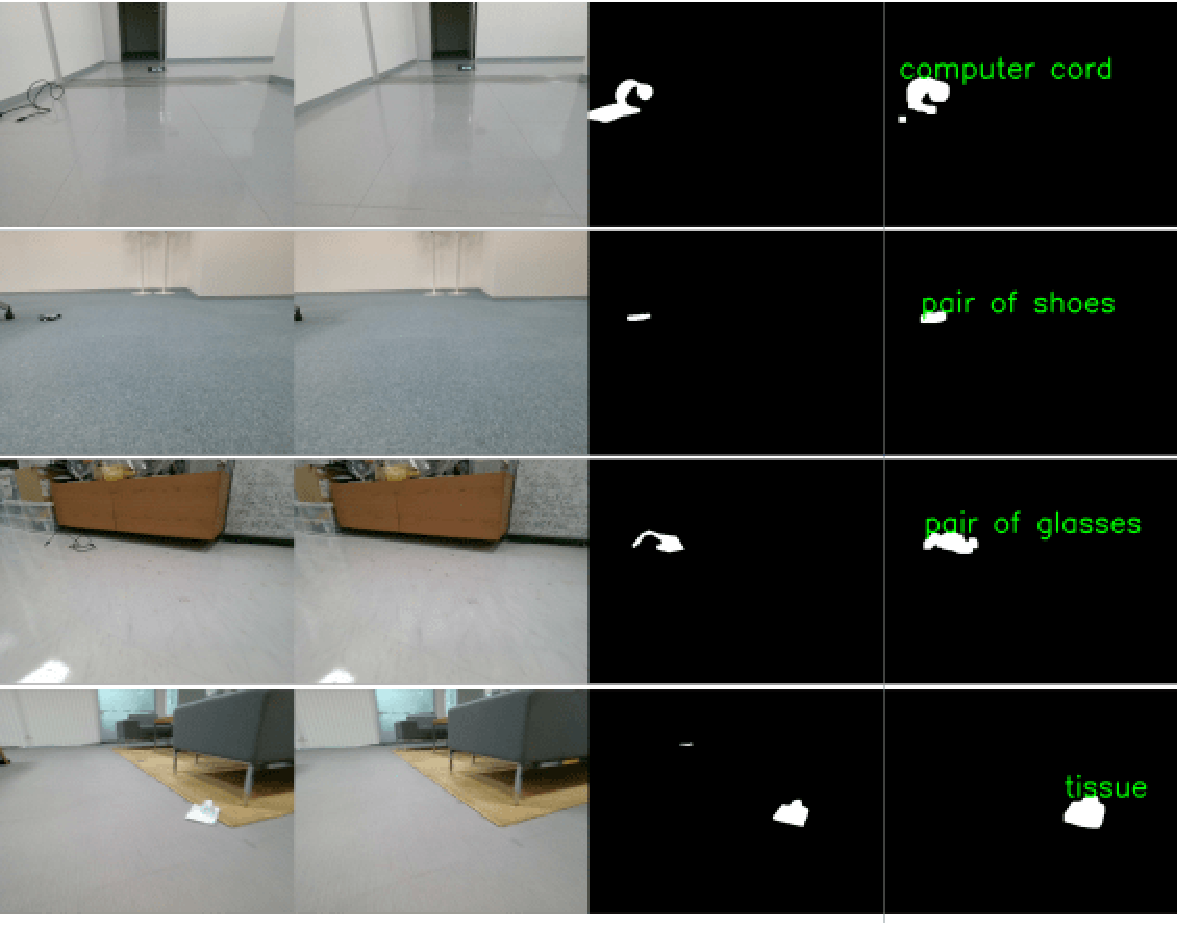}\\
\vspace*{-2mm}
  (a)~~~~~~~~~~~~~(b)~~~~~~~~~~~~~(c)~~~~~~~~~~~~~(d)
\caption{%
Results of large multimodal model. (a) live image, (b) reference image, (c) ground truth mask, (d) predicted object label }
\vspace*{-3mm}
\label{fig:C}
\end{figure}
}

\newcommand{\figD}{
\begin{figure}[t]
  \centering
  \includegraphics[width=6cm]{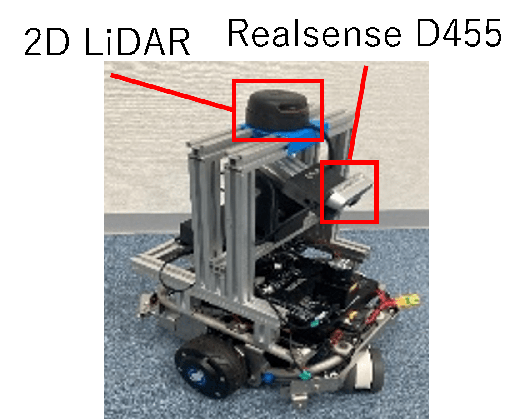}\\
\caption{%
Differential two-wheeled robot equipped with a Realsense D455, 2D LiDAR, IMU was used to collect dataset.
}
\vspace*{-6mm}
\label{fig:D}
\end{figure}
}

\newcommand{\figE}{
\begin{figure}[t]
  \centering
  \includegraphics[width=8cm]{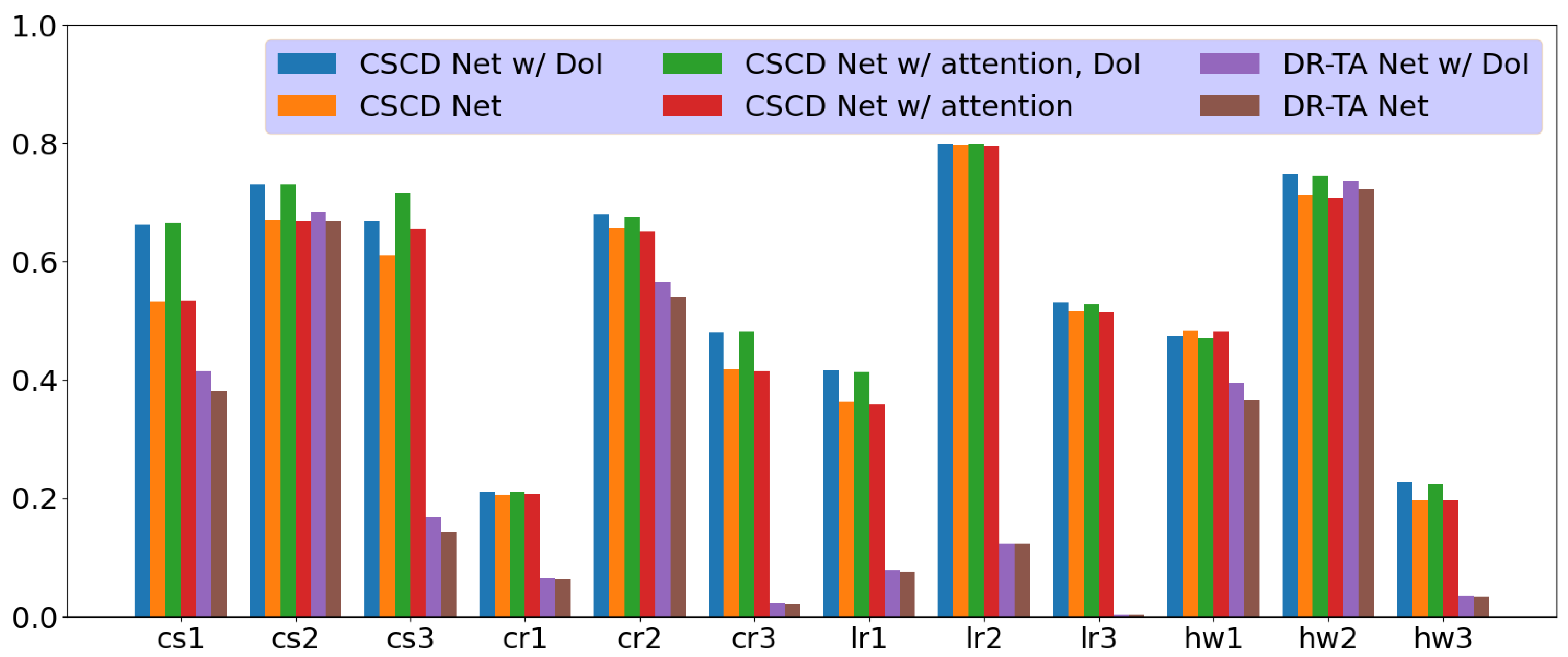}\\
\caption{%
Performance of each training-test pair. X axis indices test dataset (cs: convinience store, cr: conference room, lr: living room, hw: hall way, The number at the end represents a object group.). Y axis indices f-score of each dataset.
}
\vspace*{+6mm}
\label{fig:E}
\end{figure}
}

\newcommand{\figF}{
\begin{figure*}[t]
  \centering
  \includegraphics[width=17cm]{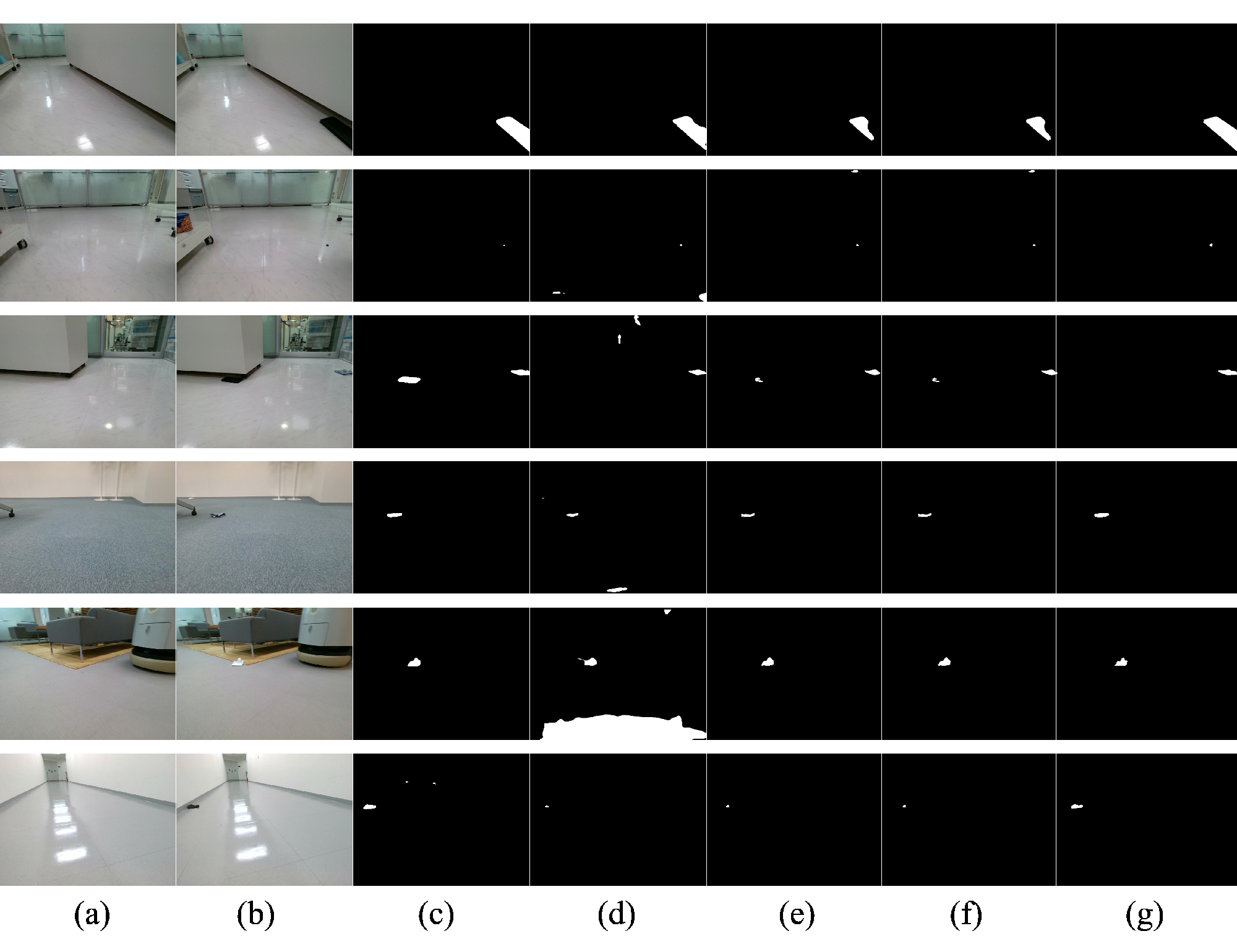}\\

  \centering
\caption{%
Qualitative results of each methods. (a) reference image, (b) live image, (c) ground truth, (d) DR-TA Net, 
(e) CSCD Net, (f) CSCD Net w/ attention and (g) Proposed method 
}
\label{fig:F}
\end{figure*}
}

\newcommand{\tabA}{
\begin{table}[t]
  \centering
  \caption{%
Mean F-Score of each methods.
}
\begin{tabular}{lc}
\toprule  Method                     & Mean F-Score \\ 
\midrule
DR-TA Net \cite{chen2021dr}                  & 0.2619                           \\ 
CSCD Net \cite{sakurada2020weakly}                   & 0.5138                           \\
CSCD Net w/ attention \cite{sakurada2020weakly,takeda2022domain}     & 0.5158                           \\
\midrule
DR-TA Net \cite{chen2021dr} w/ DoI (Ours)          & 0.2742                           \\
CSCD Net \cite{sakurada2020weakly} w/ DoI (Ours)           & 0.5524                           \\
CSCD Net w/ attention \cite{sakurada2020weakly,takeda2022domain}, w/ DoI (Ours) & {\bf 0.5550}         \\
\bottomrule                  
\end{tabular}

\label{tab:A}
\end{table}
}

\begin{abstract}
In everyday indoor navigation, robots often need to detect non-distinctive small-change objects (e.g., stationery, lost items, and junk, etc.) to maintain domain knowledge. This is most relevant to ground-view change detection (GVCD), a recently emerging research area in the field of computer vision. However, these existing techniques rely on high-quality class-specific object priors to regularize a change detector model that cannot be applied to semantically nondistinctive small objects. To address ill-posedness, in this study, we explore the concept of degree-of-ill-posedness (DoI) from the new perspective of GVCD, aiming to improve both passive and active vision. This novel DoI problem is highly domain-dependent, and manually collecting fine-grained annotated training data is expensive. To regularize this problem, we apply the concept of self-supervised learning to achieve efficient DoI estimation scheme and investigate its generalization to diverse datasets. Specifically, we tackle the challenging issue of obtaining self-supervision cues for semantically non-distinctive unseen small objects and show that novel ``oversegmentation cues" from open vocabulary semantic segmentation can be effectively exploited. When applied to diverse  real datasets, the proposed DoI model can boost state-of-the-art change detection models, and it shows stable and consistent improvements when evaluated on real-world datasets.
\end{abstract}

\section{Introduction}

In everyday indoor navigation, robots often need to detect semantically nondistinctive, small-change objects (e.g., stationery, lost items, and junk) to maintain their domain knowledge. 
In this paper, we refer to the robot's knowledge of where objects exist in the robot's workspace as domain knowledge. This knowledge is important for robots navigating through the environment.
In the field of computer vision, this is most relevant to the emerging research area of ground-view change detection (GVCD) \cite{chen2021dr,sakurada2020weakly}, which aims to train a change detector in the target domain such that it can successfully detect change objects from a single ground-view onboard camera without relying on typical change-detection devices (e.g., laser range finders, tactile sensors) or simplified vision setups (e.g., satellite-view parallel projection). This setup is highly ill-posed because of visual uncertainty combined with complex nonlinear perspective projection, and remains largely unsolved. Current methods rely on manually annotated high-quality class-specific priors of objects to regularize a change detector model (e.g., CSCD-Net \cite{sakurada2020weakly}). Thus, they are not directly applicable to non-distinctive small-change objects in robotic applications.

One of the biggest problems in semantically nondistinctive small-object change detection is the inherent ill-posedness of passive vision. Most previous studies on image change detection assumed a passive observer (i.e., a robot) and did not consider viewpoint planning and robot control. However, small object change detection can be an ill-posed problem for passive observers because, from several viewpoints, small objects may blend into the background and become indistinguishable. 
In this research, we define high degree of ill-posedness (DoI) as a state in which the robot's relative position and posture with respect to an object are not appropriate, resulting in incorrect estimation results.
Moreover, we argue that if a given visual image is ill-posed, the robot should temporarily suspend its own navigation task involving passive change detection and trigger an active change-detection task in which it approaches and closely inspects potential small object changes. Active change detection tasks should only be triggered in cases with truly high ill-posedness, because suspending the navigation task for every potential small object change requires prohibitive sensing and action costs. As a first step toward such an active change detection setup, we wish to realize a new ability for robots to estimate the ill-posedness of the passive vision task from the input visual image itself (shown in Figure \ref{fig:A}).

\figA

Predicting the DoI of small-object change detection from a given visual image is a non-trivial problem. Approaches that predict the DoI or other similar ill-posedness measures of passive vision in advance are useful for omitting costly passive vision and have been researched in the field of computer vision. For instance, in the context of image retrieval, if the DoI of an image retrieval task using a certain image feature can be known in advance of image retrieval, high-cost and low-return image retrieval can be omitted. Therefore, the task of predicting the DoI or matchability of a given visual feature has been studied \cite{Hartmann_2014_CVPR}. However, most DoI estimation tasks studied thus far assume that a fine-grained annotated training set is available for the supervised learning of the predictor. Unfortunately, in the context of small-object change detection, such assumptions often do not hold. For instance, in a typical small-object change scenario, such as lost item detection, the semantic classes of potential lost items essentially follow a long-tail distribution, and manually generating such a diverse training set with fine-grained annotations incurs prohibitive costs.

In this study, we propose a novel DoI estimation scheme for small object change detection achieved by a foundation model and a large language model, and investigated its generalization to diverse datasets. As shown in Figure \ref{fig:P}, our approach effectively leverages the rich zero-shot performance of pretrained foundation model (PFM) and large language models while performing the DoI estimation without additional training. 
In order to estimate DoI with zero shots, an object mask as accurate as ground truth is required. Therefore, we utilize the Segment Anything Model (SAM) \cite{DBLP:conf/iccv/KirillovMRMRGXW23}, which has high region segmentation capability.
Our basic assumption in this scheme is that the SAM provides exhaustive oversegmentation of all possible small objects, including potentially small change objects. SAMs excel in their ability to segment small objects with a high degree of accuracy. Although such a thorough oversegmentation may naturally contain many false positives, it is still valuable in evaluating the consensus between the change detector and the oversegmentation. When evaluated on realistic real-world datasets, the proposed DoI model can boost SOTA change-detection models and shows stable and consistent improvements.

\section{Related Work}

Image change detection is the problem of detecting changes between two scenes and is used for anomaly detection in fixed surveillance cameras \cite{7795955} and farmland analysis from satellite images \cite{9298743}. Common problem settings for change detection is top-down-view change detection \cite{8444434,8618401}. However, this simple top-down vision setup often relies on the assumption that pixel-to-pixel correspondences are simplified. In contrast, for GVCD, the problem is highly ill-posed because of complex perspective projection mapping.

Siamese-based convolutional neural networks (CNNs) are commonly used for ground view change detection. Sakurada et al. verified the effectiveness of CNNs in change-detection applications \cite{DBLP:conf/bmvc/SakuradaO15}. Subsequently, a method using optical flow for image registration was proposed; however, this method has a high computational cost \cite{sakurada2017dense}. Therefore, CSCD-Net \cite{sakurada2020weakly} using a correlation layer \cite{dosovitskiy2015flownet} was proposed, and its effectiveness was demonstrated on public PCD datasets \cite{DBLP:conf/bmvc/SakuradaO15}. Furthermore, the DR-TA \cite{chen2021dr} Net was proposed as a method to reduce the computational cost and achieved state of the art performance.
Several studies have been proposed recently that focus on small object changes \cite{takeda2022domain,takeda2023lifelong,klomp2019real}, although all have only been considered in a passive vision context.


Our approach is also related to the pretrained foundation model (PFM). PFM was trained using large-scale internet-scale data \cite{zhou2023comprehensive}. Because it is trained on large amounts of data, it is highly generalizable and achieves high zero-shot ability in many fields. In natural language, this applies to Large Language Models (LLMs) such as BERT \cite{devlin2019bert}. In computer vision, an SAM \cite{DBLP:conf/iccv/KirillovMRMRGXW23} that can extract object regions specified by points, text, or bounding boxes. Furthermore, multimodally extended models have been proposed, and representative examples are CLIP \cite{radford2021learning}, ALIGN \cite{jia2021scaling}, and LLaVA \cite{liu2023improved}. CLIP and ALIGN are instances of vision-language models (VLMs) that use contrastive learning to train features obtained from images and those obtained from language modalities such that similar features are closer together and different features are further apart. Furthermore, a technology known as Alpha CLIP \cite{sun2023alpha} has been proposed, which is a derivative of this technology and allows for feature extraction specific to the region of interest by specifying the region of interest using an alpha mask. 
Furthermore, Large language-and-vision assistant (LLaVA) \cite{liu2023improved} has been proposed as a type of large-scale multimodal model.
LLaVA uses a two-stage instruction-tuning procedure to connect the pretrained CLIP ViT-L/336px visual encoder and the large language model Vicuna \cite{vicuna2023} to perform VQA and achieves SOTA performance in 12 benchmarks including VQA.

However, the use of PFM for change detection remains largely unexplored. For instance, \cite{obinata2023semantic} was a previous study that utilized LLMs for semantic scene difference detection in daily life by mobile robots using a PFM. To the best of our knowledge, this is the first study to utilize PFM for semantically nondistinctive change-detection tasks.

\section{Approach}
\figP

To realize a change-detection framework utilizing DoI, the change detector and PFM interacted effectively in the proposed framework. Figure \ref{fig:P} shows proposed change detection framework. The framework consists of three modules: base change detection, object search, and integration modules. The framework takes two images as inputs: a live image currently acquired by the robot and a reference image acquired by the robot from the same viewpoint as that in the past.
First, the base change-detection module outputs a binary mask, where a pixel value of 1 represents a changed part and that of 0 represents an unchanged part.
Next, the object search module extracts the object language information using the Large Multimodal Model (MLL) and performs Open Vocabulary Segmentation (OVS) using the SAM.
Finally, the integration module integrates the base change detection results and OVS results while evaluating the pixel-wise DoI. Thus, the change detection results were obtained.

\subsection{Base Change Detection Module}

Following the literature on image change detection, change detection is formulated as a two-input, one-output image recognition task that takes a paired reference and live image as input, which is a color image with RGB pixel values, and outputs a change mask of pixel-wise probability values. We selected CSCD Net \cite{sakurada2020weakly} with siamese attention mask \cite{takeda2022domain} as the basic change detector because its Siamese structure architecture with correlation layers is effective in ground-view change-detection applications among existing Siamese structure change detectors. And also siamese attention mask is expected to alleviate false positives of non object region such as shadow and lighting reflection.

\figB

Unlike typical image change detection applications such as satellite imagery \cite{9298743} or three-dimensional (3D) building modeling \cite{taneja2013city}, for small object change detection, the cost of manually preparing fine-grained annotations is prohibitive, and a complete training dataset is not available. Therefore, we adopt the generation of a pseudo-training set using small objects generated based on the COCO dataset \cite{lin2014microsoft} as examples of small change objects, as in \cite{takeda2023lifelong}. Specifically, a synthetic change dataset in which foreground objects from the COCO dataset were pasted onto the background of a reference image was used as a pseudo-training set. The COCO dataset provides a diverse sample of frequently occurring objects in indoor scenes (e.g., stationery, lost items, and junks) with ground-truth annotations. Specifically, training samples were generated using the following procedure: First, one of the images selected from the reference images was sampled and used as a training background image. Next, the COCO object was sampled and pasted to a random size at a random position in the training background image. The resulting synthetic images were used as the training samples. Figure \ref{fig:B} shows an overview of the training set creation task.
\subsection{Object Search Module}
The detection of object changes in a real-world 3D space from two-dimensional (2D) images is a significantly ill-posed problem. A straightforward method to address this challenge is to exploit a priori knowledge to regularize the problem \cite{sakurada2020weakly}. Semantic information is a promising type of a priori knowledge. For instance, Siamese-structured neural networks, which are commonly used for change detection, use weights as a backbone that are pretrained on large datasets, such as the ImageNet dataset \cite{deng2009imagenet}. Thus, if the domain is similar to ImageNet, regularization can be performed using the semantic information learned from ImageNet as prior knowledge. However, it is not applicable for detecting changes in small, semantically non-distinctive objects. Therefore, we are interested in developing new regularization methods for small object change detection.


To address this issue, 
we propose to utilize the high zero-shot capability of PFM \cite{zhou2023comprehensive} and the feature consistency of language and images of lLLMs.
The PFM is pretrained with large-scale data that are almost infinitely available on the Internet, and the PFM can be directly applied to downstream visual recognition tasks without fine-tuning. 
Also, LMMs, especially that can take image and language as input, are trained to match image features and language features by large scale image-text pair. This allows for an almost infinite number of vocabulary words to be associated with images. This characteristic might be valid for change detection where the detection target object class follows a long-tail distribution.


Our approach consist of linguistic information extraction and OVS. For linguistic information extraction, 
we used a LLaVA1.5 \cite{liu2023improved} framework using Alpha CLIP \cite{sun2023alpha} 
The Alpha CLIP was employed because it can enhance the features of masked areas and may emphasize the features of small objects. LLaVA1.5 was also selected because it achieves state-of-the-art performance among existing large-scale multimodal models.
For object mask generation, we employ a text-promptable segmentation method using the SAM \cite{DBLP:conf/iccv/KirillovMRMRGXW23} for segmentation and the Grounding DINO zero-shot object detection model \cite{liu2023grounding}.
This method was chosen because it allows the specification of regions using language as input and because the use of SAM allows high region segmentation even for small objects.

\figC

The object search module is triggered by the acquisition of change-detection results using the base change-detection module. First, we obtain language information for the target domain by inputting the live image, change detection results, and query statement into the LMM. In this study, we adopted 
\begin{itemize}
        \item
        {\it ``What is the class name of this object? Please answer like 'This object is ..." }
        \end{itemize}
as the prompt sentence to obtain linguistic information about object labels. The results are shown in Figure \ref{fig:C}. We observe cases in which the change-detection results do not accurately capture the object region. To address this issue, the change detection results from the base change detection module are modified by applying a dilation operation several times with a kernel size of 5$\times$5. Here, the language information of the object name can be obtained. However, it contains a large amount of noise. Therefore, to remove the evident noise, the language labels containing the word 
{\it ``floor"}
are removed. Subsequently, object region extraction is performed using the acquired language information. Specifically, OVS is performed using reference and live images based on the obtained linguistic information. For OVS, we use implementation of lang-segment-anything\footnote{https://github.com/luca-medeiros/lang-segment-anything} that uses Grounding-DINO cite{liu2023grounding} and SAM \cite{DBLP:conf/iccv/KirillovMRMRGXW23} and specify a language label for extracting the object region. In this manner, the object regions of the reference $O_r$ and live images $O_l$ are acquired.

Furthermore, we develop a novel application that uses this property to predict the object names of small detected object changes. This application is based on another desirable property of our method; it is based on an LLM and can predict the meanings of object segments. Such an application is useful for lost-and-found detection by listing detected objects and linking them to object names.

\subsection{Integration Module}

The integration module integrates the results of open vocabulary object segmentation for live image $O_l$ and reference image $O_r$ and base change detection results $M_o$, evaluates the DoI, and gets binary change mask.
We estimate the DoI as follows:

\begin{equation}
        DoI = f_{b}{(O_l, O_r)} \times (1 - f_{IoU}{(O_l, M_o)})
\end{equation}

where, $f_{b}$ is binary function that return 1 if given two binary mask does not have overlap. And $f_{IoU}$ is function that calculate Intersection over Union between two binary masks.
If the DoI is greater than 0 and smaller than 0.9, then the $O_l$ is more reliable than original change detection result. And the over-segmentation mask $O_l$ is adopted as the final change detection result. Otherwise $M_o$ is adopted as the final change detection result.






\section{Experiments}
\label{sec:A}

\subsection{Dataset}
\figD

To verify whether the performance of the proposed method can be generalized to real-world workspaces, we conducted real-world experiments using a robot developed in our laboratory. Figure \ref{fig:D} shows an overview of the robot. The robot was a differential two-wheeled robot equipped with a Realsense D455 camera, 2D LiDAR, and IMU sensors. The resolution of image is set to $640 \times 480$. The workspace environment includes convinience store, conference rooms, living rooms, and hallways. An everyday navigation scenario was assumed in which a robot was engaged in a point-goal navigation task (e.g., delivery task). Specifically, a image sequence was acquired during point-goal navigation while running along a wall using a joystick and performing self-localization using 2D laser SLAM. 
Three navigation task sessions were performed. One is the 2D LIDAR SLAM task, which aims to construct an occupancy grid map that is used to locate viewpoints in the following two tasks. The second is the task of collecting reference images. The third is the change detection task.
To obtain live and reference image pairs that are input into the change detector, each live image is paired with the reference image with the closest viewpoint in the 3DoF viewpoint space. To simulate a typical small-object change-detection scenario of lost item detection, groups of small objects were placed at random locations in the environment. To investigate the dependence of performance on the small object groups, three small object groups were created. Group 1: Smartphones, cables, notebooks, and pens. Group 2: Handkerchiefs, wallets, and IC cards. Group 3: Screws, batteries, clips. A total of 12 datasets were created by combining four workspaces and three small object groups. The number of images was 1875 and from 82 to 314 for each group. The annotation was created in the form of a binary change mask using PaddleSeg \cite{liu2021paddleseg,paddleseg2019} as input assistance.

\subsection{Settings}

Because the reference and live images are not acquired at exactly the same location, pixel-level misalignment errors can occur, which can degrade change detection performance. To address this issue, we first performed a pixel-wise alignment using PDC Net \cite{truong2021learning}. NVIDIA RTX 3090 was used to train the base change detector CSCD Net and DR-TA Net. The backbone for both CSCD Net and DR-TA Net was ResNet-18 \cite{he2016deep}. For CSCD Net, the learning rate was 0.0001, the optimization method was Adam \cite{KingBa15}, the batch size was 32, and weights after 40,000 iteration was used.  For DR-TA Net, the learning rate was 0.001, the optimization method was Adam \cite{KingBa15}, the batch size was 16, and weights after 15,000 epoch was used. For inference of change detector,  probability value over 0.5 was treated as changed pixels.


\subsection{Comparing Methods}

DR-TA Net \cite{chen2021dr}, CSCD Net \cite{sakurada2020weakly}, and CSCD Met w/ Siamese attention mask \cite{takeda2022domain} were used for comparison. DR-TA Net is change detection method that combines dynamic receptive fields and temporal attention, following the implementation of \cite{chen2021dr}. CSCD Net is a method presented in \cite{sakurada2020weakly} that addresses misalignment errors between reference images and live images using a correlation layer imported from the field of visual tracking. The Siamese attention mask is a method developed in our previous study that combines features used in visual place recognition with the middle layer of a Siamese encoder. 
As in previous study \cite{takeda2022domain}, performance was evaluated by F-Score calculated by change detection result and ground truth change mask.
F-Score is a measure of the coincidence between the change detection result and the ground truth mask and is defined by the following formula:

\begin{equation}
        F-Score = \frac{2 \times Precision \times Recall}{Precision + Recall} 
\end{equation}

\begin{equation}
        Precision = \frac{TP}{TP + FP}
\end{equation}

\begin{equation}
        Recall = \frac{TP}{TP + FN}
\end{equation}

Let TP be the total number of True Positive pixels, TN the total number of True Negative pixels, FP the total number of False Positive pixels, and FN the total number of False Negative pixels.


\subsection{Result}
\figE
\tabA

\figF

Figure \ref{fig:E} shows the performance of each training-test pair. This result confirms that the proposed method has a significantly better performance than the conventional method. In particular, the performance improvement rate is high for the convinience store dataset. 
Convenience store is cluttered environments with many objects, and the DoI is often high due to the cluttered environment. In such an environment, our framework accurately estimated the DoI while making good use of over-segmentation, resulting in improved performance.
However, some datasets (cr1, lr2) showed little performance improvement. This is due to the large number of false negatives in the base change detector, indicating that even small areas do not perform well when changes are not detected. DoI estimation in undetected regions is an important direction for our future research.

The average F-Score with and without applying the DoI estimation method to all conventional methods is shown in Table \ref{tab:A}.
For all methods, we observed improved performance when the DoI estimation method was applied. 
This confirms that our proposed system is effective regardless of the change detection algorithm.
This is because the high zero-shot inference capability of the PFM, which allows the change detector to generate over-segmentation masks that can be used for DoI prediction in both cases with many false positives and false negatives, thus improving performance regardless of the accuracy of the change detector.

Figure \ref{fig:F} shows comparison of output results between our proposed method and conventional methods. Even in cases where the original change detection method did not recognize the object region well, the proposed method successfully detected it, thereby improving the change detection performance. Our proposed method can accurately detect irregularly shaped objects, such as handkerchief, and objects that are hardly identified on the floor, such as notebooks. This is attributed to the successful use of the zero-shot inference capability of the LLM and the underlying model, which enabled detection.






\section{Conclusion}
In this paper, we explored the concept of degree-of-ill-posedness (DoI) from the new perspective of small object change detection, aiming to improve both passive and active vision. To this end, we proposed novel zero-shot DoI estimation scheme incorporating pretrained foundation model. Extensive experiments have shown that our DoI estimation scheme is effective. This study focused on leveraging DoI for the purpose of evaluating how reliable the output of a passive change detector is. Our immediate future work includes extending the framework from passive to active change detectors by using the DoI to determine the appropriate next-best-view.

\bibliography{ref}

\begin{thebibliography}{10}

\bibitem{chen2021dr}
Shuo Chen, Kailun Yang, and Rainer Stiefelhagen.
\newblock Dr-tanet: Dynamic receptive temporal attention network for street
  scene change detection.
\newblock In {\em 2021 IEEE Intelligent Vehicles Symposium (IV)}, pages
  502--509. IEEE, 2021.

\bibitem{sakurada2020weakly}
Ken Sakurada, Mikiya Shibuya, and Weimin Wang.
\newblock Weakly supervised silhouette-based semantic scene change detection.
\newblock In {\em 2020 IEEE International conference on robotics and automation
  (ICRA)}, pages 6861--6867. IEEE, 2020.

\bibitem{Hartmann_2014_CVPR}
Wilfried Hartmann, Michal Havlena, and Konrad Schindler.
\newblock Predicting matchability.
\newblock In {\em Proceedings of the IEEE Conference on Computer Vision and
  Pattern Recognition (CVPR)}, June 2014.

\bibitem{DBLP:conf/iccv/KirillovMRMRGXW23}
Alexander Kirillov, Eric Mintun, Nikhila Ravi, Hanzi Mao, Chlo{\'{e}} Rolland,
  Laura Gustafson, Tete Xiao, Spencer Whitehead, Alexander~C. Berg, Wan{-}Yen
  Lo, Piotr Doll{\'{a}}r, and Ross~B. Girshick.
\newblock Segment anything.
\newblock In {\em {IEEE/CVF} International Conference on Computer Vision,
  {ICCV} 2023, Paris, France, October 1-6, 2023}, pages 3992--4003. {IEEE},
  2023.

\bibitem{7795955}
Matthias Michael, Christian Feist, Florian Schuller, and Marc Tschentscher.
\newblock Fast change detection for camera-based surveillance systems.
\newblock In {\em 2016 IEEE 19th International Conference on Intelligent
  Transportation Systems (ITSC)}, pages 2481--2486, 2016.

\bibitem{9298743}
Ran Jing, Zhaoning Gong, and Hongliang Guan.
\newblock Land cover change detection with vhr satellite imagery based on
  multi-scale slic-cnn and scae features.
\newblock {\em IEEE Access}, 8:228070--228087, 2020.

\bibitem{8444434}
Shunping Ji, Shiqing Wei, and Meng Lu.
\newblock Fully convolutional networks for multisource building extraction from
  an open aerial and satellite imagery data set.
\newblock {\em IEEE Transactions on Geoscience and Remote Sensing},
  57(1):574--586, 2019.

\bibitem{8618401}
Tao Lei, Yuxiao Zhang, Zhiyong Lv, Shuying Li, Shigang Liu, and Asoke~K. Nandi.
\newblock Landslide inventory mapping from bitemporal images using deep
  convolutional neural networks.
\newblock {\em IEEE Geoscience and Remote Sensing Letters}, 16(6):982--986,
  2019.

\bibitem{DBLP:conf/bmvc/SakuradaO15}
Ken Sakurada and Takayuki Okatani.
\newblock Change detection from a street image pair using {CNN} features and
  superpixel segmentation.
\newblock In Xianghua Xie, Mark~W. Jones, and Gary K.~L. Tam, editors, {\em
  Proceedings of the British Machine Vision Conference 2015, {BMVC} 2015,
  Swansea, UK, September 7-10, 2015}, pages 61.1--61.12. {BMVA} Press, 2015.

\bibitem{sakurada2017dense}
Ken Sakurada, Weimin Wang, Nobuo Kawaguchi, and Ryosuke Nakamura.
\newblock Dense optical flow based change detection network robust to
  difference of camera viewpoints.
\newblock {\em arXiv preprint arXiv:1712.02941}, 2017.

\bibitem{dosovitskiy2015flownet}
Alexey Dosovitskiy, Philipp Fischer, Eddy Ilg, Philip Hausser, Caner Hazirbas,
  Vladimir Golkov, Patrick Van Der~Smagt, Daniel Cremers, and Thomas Brox.
\newblock Flownet: Learning optical flow with convolutional networks.
\newblock In {\em Proceedings of the IEEE international conference on computer
  vision}, pages 2758--2766, 2015.

\bibitem{takeda2022domain}
Koji Takeda, Kanji Tanaka, and Yoshimasa Nakamura.
\newblock Domain invariant siamese attention mask for small object change
  detection via everyday indoor robot navigation.
\newblock In {\em 2022 IEEE/RSJ International Conference on Intelligent Robots
  and Systems (IROS)}, pages 739--745. IEEE, 2022.

\bibitem{takeda2023lifelong}
Koji Takeda, Kanji Tanaka, and Yoshimasa Nakamura.
\newblock Lifelong change detection: Continuous domain adaptation for small
  object change detection in everyday robot navigation.
\newblock In {\em 2023 18th International Conference on Machine Vision and
  Applications (MVA)}, pages 1--5. IEEE, 2023.

\bibitem{klomp2019real}
Sander~R Klomp, Dennis~WJM van~de Wouw, et~al.
\newblock Real-time small-object change detection from ground vehicles using a
  siamese convolutional neural network.
\newblock {\em Journal of Imaging Science and Technology}, 63(6):060402, 2019.

\bibitem{zhou2023comprehensive}
Ce~Zhou, Qian Li, Chen Li, Jun Yu, Yixin Liu, Guangjing Wang, Kai Zhang, Cheng
  Ji, Qiben Yan, Lifang He, Hao Peng, Jianxin Li, Jia Wu, Ziwei Liu, Pengtao
  Xie, Caiming Xiong, Jian Pei, Philip~S. Yu, and Lichao Sun.
\newblock A comprehensive survey on pretrained foundation models: A history
  from bert to chatgpt, 2023.

\bibitem{devlin2019bert}
Jacob Devlin, Ming-Wei Chang, Kenton Lee, and Kristina Toutanova.
\newblock Bert: Pre-training of deep bidirectional transformers for language
  understanding, 2019.

\bibitem{radford2021learning}
Alec Radford, Jong~Wook Kim, Chris Hallacy, Aditya Ramesh, Gabriel Goh,
  Sandhini Agarwal, Girish Sastry, Amanda Askell, Pamela Mishkin, Jack Clark,
  et~al.
\newblock Learning transferable visual models from natural language
  supervision.
\newblock In {\em International conference on machine learning}, pages
  8748--8763. PMLR, 2021.

\bibitem{jia2021scaling}
Chao Jia, Yinfei Yang, Ye~Xia, Yi-Ting Chen, Zarana Parekh, Hieu Pham, Quoc Le,
  Yun-Hsuan Sung, Zhen Li, and Tom Duerig.
\newblock Scaling up visual and vision-language representation learning with
  noisy text supervision.
\newblock In {\em International conference on machine learning}, pages
  4904--4916. PMLR, 2021.

\bibitem{liu2023improved}
Haotian Liu, Chunyuan Li, Yuheng Li, and Yong~Jae Lee.
\newblock Improved baselines with visual instruction tuning.
\newblock {\em arXiv preprint arXiv:2310.03744}, 2023.

\bibitem{sun2023alpha}
Zeyi Sun, Ye~Fang, Tong Wu, Pan Zhang, Yuhang Zang, Shu Kong, Yuanjun Xiong,
  Dahua Lin, and Jiaqi Wang.
\newblock Alpha-clip: A clip model focusing on wherever you want.
\newblock {\em arXiv preprint arXiv:2312.03818}, 2023.

\bibitem{vicuna2023}
Wei-Lin Chiang, Zhuohan Li, Zi~Lin, Ying Sheng, Zhanghao Wu, Hao Zhang, Lianmin
  Zheng, Siyuan Zhuang, Yonghao Zhuang, Joseph~E. Gonzalez, Ion Stoica, and
  Eric~P. Xing.
\newblock Vicuna: An open-source chatbot impressing gpt-4 with 90\%* chatgpt
  quality, March 2023.

\bibitem{obinata2023semantic}
Yoshiki Obinata, Kento Kawaharazuka, Naoaki Kanazawa, Naoya Yamaguchi, Naoto
  Tsukamoto, Iori Yanokura, Shingo Kitagawa, Koki Shinjo, Kei Okada, and
  Masayuki Inaba.
\newblock Semantic scene difference detection in daily life patroling by mobile
  robots using pre-trained large-scale vision-language model.
\newblock In {\em 2023 IEEE/RSJ International Conference on Intelligent Robots
  and Systems (IROS)}, pages 3228--3233. IEEE, 2023.

\bibitem{taneja2013city}
Aparna Taneja, Luca Ballan, and Marc Pollefeys.
\newblock City-scale change detection in cadastral 3d models using images.
\newblock In {\em Proceedings of the IEEE Conference on computer Vision and
  Pattern Recognition}, pages 113--120, 2013.

\bibitem{lin2014microsoft}
Tsung-Yi Lin, Michael Maire, Serge Belongie, James Hays, Pietro Perona, Deva
  Ramanan, Piotr Doll{\'a}r, and C~Lawrence Zitnick.
\newblock Microsoft coco: Common objects in context.
\newblock In {\em Computer Vision--ECCV 2014: 13th European Conference, Zurich,
  Switzerland, September 6-12, 2014, Proceedings, Part V 13}, pages 740--755.
  Springer, 2014.

\bibitem{deng2009imagenet}
Jia Deng, Wei Dong, Richard Socher, Li-Jia Li, Kai Li, and Li~Fei-Fei.
\newblock Imagenet: A large-scale hierarchical image database.
\newblock In {\em 2009 IEEE conference on computer vision and pattern
  recognition}, pages 248--255. Ieee, 2009.

\bibitem{liu2023grounding}
Shilong Liu, Zhaoyang Zeng, Tianhe Ren, Feng Li, Hao Zhang, Jie Yang, Chunyuan
  Li, Jianwei Yang, Hang Su, Jun Zhu, et~al.
\newblock Grounding dino: Marrying dino with grounded pre-training for open-set
  object detection.
\newblock {\em arXiv preprint arXiv:2303.05499}, 2023.

\bibitem{liu2021paddleseg}
Yi~Liu, Lutao Chu, Guowei Chen, Zewu Wu, Zeyu Chen, Baohua Lai, and Yuying Hao.
\newblock Paddleseg: A high-efficient development toolkit for image
  segmentation.
\newblock {\em arXiv preprint arXiv:2101.06175}, 2021.

\bibitem{paddleseg2019}
PaddlePaddle Authors.
\newblock Paddleseg, end-to-end image segmentation kit based on paddlepaddle.
\newblock \url{https://github.com/PaddlePaddle/PaddleSeg}, 2019.

\bibitem{truong2021learning}
Prune Truong, Martin Danelljan, Luc Van~Gool, and Radu Timofte.
\newblock Learning accurate dense correspondences and when to trust them.
\newblock In {\em Proceedings of the IEEE/CVF Conference on Computer Vision and
  Pattern Recognition}, pages 5714--5724, 2021.

\bibitem{he2016deep}
Kaiming He, Xiangyu Zhang, Shaoqing Ren, and Jian Sun.
\newblock Deep residual learning for image recognition.
\newblock In {\em Proceedings of the IEEE conference on computer vision and
  pattern recognition}, pages 770--778, 2016.

\bibitem{KingBa15}
Diederik Kingma and Jimmy Ba.
\newblock Adam: A method for stochastic optimization.
\newblock In {\em International Conference on Learning Representations (ICLR)},
  San Diega, CA, USA, 2015.

\end{thebibliography}
\bibliographystyle{unsrt}

\end{document}